\documentclass[10pt,letterpaper]{article}

\usepackage[top=0.85in,left=1.5in,footskip=0.75in]{geometry}
\usepackage[numbers]{natbib}
\usepackage[affil-it]{authblk}
\usepackage{float}

\usepackage{url,microtype,subcaption}
\usepackage[onehalfspacing]{setspace}
\usepackage{caption}
\usepackage{makecell}
\usepackage[edges]{forest}
\usepackage{multirow}
\usepackage{graphicx}
\usepackage{fancyhdr}
\fancypagestyle{firstpage}{%
  \lhead{\textcolor{red}{This manuscript is published in Frontiers in Neuroinformatics. Please cite it as:} \\
\textcolor{blue}{Hazan H, Saunders DJ, Khan H, Patel D, Sanghavi DT, Siegelmann HT and Kozma R (2018) 
\textit{BindsNET: A Machine Learning-Oriented Spiking Neural Networks Library in Python.}
Front. Neuroinform. 12:89. doi: 10.3389/fninf.2018.00089}}
  \rhead{ }
}

\newcommand\blfootnote[1]{
  \begingroup
  \renewcommand\thefootnote{}\footnote{#1}
  \addtocounter{footnote}{-1}
  \endgroup
}

\onecolumn

\title{BindsNET: A machine learning-oriented spiking neural networks library in Python}

\author{Hananel Hazan}
\author{Daniel J. Saunders\thanks{Corresponding author (\texttt{djsaunde@cs.umass.edu})}}
\author{Hassaan Khan}
\author{Darpan T. Sanghavi}
\author{Hava T. Siegelmann}
\author{Robert Kozma}

\affil{Biologically Inspired Neural and Dynamical Systems Laboratory, University of Massachusetts Amherst, College of Computer and Information Sciences, Amherst, MA, USA}

\begin{document}

\maketitle
\thispagestyle{firstpage}

\begin{abstract}

The development of spiking neural network simulation software is a critical component enabling the modeling of neural systems and the development of biologically inspired algorithms. Existing software frameworks support a wide range of neural functionality, software abstraction levels, and hardware devices, yet are typically not suitable for rapid prototyping or application to problems in the domain of machine learning. In this paper, we describe a new Python package for the simulation of spiking neural networks, specifically geared towards machine learning and reinforcement learning. Our software, called \texttt{BindsNET}, enables rapid building and simulation of spiking networks and features user-friendly, concise syntax. \texttt{BindsNET} is built on top of the \texttt{PyTorch} deep neural networks library, enabling fast CPU and GPU computation for large spiking networks. The \texttt{BindsNET} framework can be adjusted to meet the needs of other existing computing and hardware environments, e.g., \texttt{TensorFlow}. We also provide an interface into the OpenAI \texttt{gym} library, allowing for training and evaluation of spiking networks on reinforcement learning problems. We argue that this package facilitates the use of spiking networks for large-scale machine learning experimentation, and show some simple examples of how we envision \texttt{BindsNET} can be used in practice. \blfootnote{\texttt{BindsNET} code is available at \texttt{https://github.com/Hananel-Hazan/bindsnet}}


\end{abstract}

\section{Introduction}

The recent success of deep learning models in computer vision, natural language processing, and other domains \citep{y._lecun_deep_2015} have led to a proliferation of machine learning software packages \citep{paszke2017automatic, tensorflow2015-whitepaper, Chen2015MXNetAF, 2016arXiv160502688full, jia2014caffe, chainer_learningsys2015}. GPU acceleration of deep learning primitives has been a major proponent of this success \citep{Chetlur2014cuDNNEP}, as their massively parallel operation enables rapid processing of layers of independent nodes. Since the biological plausibility of deep neural networks is often disputed \citep{backprop_biological}, interest in integrating the algorithms of deep learning with long-studied ideas in neuroscience has been mounting \citep{10.3389/fncom.2016.00094}, both as a means to increase machine learning performance and to better model learning and decision-making in biological brains \citep{Wang295964}.

Spiking neural networks (SNNs) \citep{maass_lower_1996, maass_networks_1997, w._gerstner_spiking_2002} are sometimes referred to as the ``third generation'' of neural networks because of their potential to supersede deep learning methods in the fields of computational neuroscience \citep{10.3389/fncom.2013.00182} and biologically plausible machine learning (ML) \citep{BengioLBL15}. SNNs are also thought to be more practical for data-processing tasks in which the data has a temporal component since the neurons which comprise SNNs naturally integrate their inputs over time. Moreover, their binary (spiking or no spiking) operation lends itself well to fast and energy efficient simulation on hardware devices.

Although spiking neural networks are not widely used as machine learning systems, recent work shows that they have the potential to be. SNNs are often trained with unsupervised learning rules to learn a useful representation of a dataset, which are then used as features for supervised learning methods \citep{10.3389/fncom.2018.00024, diehl_fast-classifying_2015, hazanetal2018, saundersetal18}. Trained deep neural networks may be converted to SNNs and implemented in hardware while maintaining good image recognition performance \citep{diehl_fast-classifying_2015}, demonstrating that SNNs can in principle compete with deep learning methods. In similar lines of work \citep{snn_backprop, Hunsberger2015SpikingDN, OConnor2016DeepSN}, the popular back-propagation algorithm has been applied to differentiable versions of SNNs to achieve competitive performance on standard image classification benchmarks, more evidence in support of the potential of spiking networks for ML problem solving.

The membrane potential (or voltage) of a spiking neuron is often described by ordinary differential equations. The membrane potential of the neuron is increased or decreased by \emph{presynaptic} inputs, depending on their sign and strength. In the case of the leaky integrate-and-fire (LIF) model \citep{w._gerstner_spiking_2002} and several other models, the neuron is always decaying to a \emph{rest potential} $v_{rest}$. When a neuron reaches its \emph{threshold voltage} $v_{thr}$, it emits a spike which travels to downstream neurons via synapses, its \emph{postsynaptic} effect modulated by synaptic strengths, and its voltage is reset to some $v_{reset}$. Connections between neurons may also have their own dynamics, which can be modified by prescribed learning rules or external reward signals.

Several software packages for the discrete-time simulation of SNNs exist, with varying levels of biological realism and support for hardware platforms. Many such solutions, however, were not developed to target ML applications, and often feature abstruse syntax resulting in steep learning curves for new users. Moreover, packages with a large degree of biological realism may not be appropriate for problems in ML, since they are computationally expensive to simulate and may require a large degree of hyper-parameter tuning. Real-time hardware implementations of SNNs exist as well, but cannot support the rapid prototyping that some software solutions can.

Motivated by the foregoing shortcomings, we present the \texttt{BindsNET} spiking neural networks library, which is developed on top of the popular \texttt{PyTorch} deep learning library \citep{paszke2017automatic}. At its core, the software allows users to build, train, and evaluate SNNs composed of groups of neurons and their connections. The learning of connection weights is supported by various algorithms from the biological learning literature \citep{hebb-organization-of-behavior-1949, Markram1997RegulationOS}. A separate module provides an interface to the OpenAI \texttt{gym} \citep{BrockmanCPSSTZ16} reinforcement learning (RL) environments library from \texttt{BindsNET}. A \texttt{Pipeline} object is used to streamline the interaction between spiking networks and RL environments, removing many of the messy details from the purview of the experimenter. Still other modules provide functions such as loading of ML datasets, encoding of raw data into spike train network inputs, plotting of network state variables and outputs, and evaluation of SNN as ML models.

The paper is structured as follows: we begin in Section \ref{s:review} with an assessment of the existing SNN simulation software and hardware implementations. In Section \ref{s:structure}, the \texttt{BindsNET} library is described at a high level of detail, emphasizing the functional role of each software module, how they were motivated, and how they inter-operate. Code snippets and simple case studies are presented in Section \ref{s:examples} to demonstrate the breadth of possible \texttt{BindsNET} applications. Desirable future features are listed in \ref{s:future}, and potential research impacts are assessed in Section \ref{s:discussion}.

\section{Review}
\label{s:review}

In the last two decades, neural networks have become increasingly prominent in machine learning and artificial intelligence research, leading to a proliferation of efficient software packages for their training, evaluation, and deployment. On the other hand, the simulation of the ``third generation'' of neural networks has not been able to reach its full potential, due in part to their inherent complexity and computational requirements. However, spiking neurons excel at remembering a short-term history of their activation and feature efficient binary communication with other neurons, a useful feature in reducing energy requirements on neuromorphic hardware. Spiking neurons exhibit more properties from their biological counterpart than the computing units utilized by deep neural networks, which may constitute an important advantage in terms of practical computational power or ML performance.

Researchers that want to conduct experiments with networks of spiking neurons for ML purposes have a number of options for SNN simulation software. Many frameworks exist, but each is tailored toward specific application domains. In this section, we describe the existing relevant software libraries and the challenges associated with each, and contrast these with the strengths of our package.

We believe that the chosen simulation framework must be easy to develop in, debug, and run, and, most importantly, support the level of biological complexity desired by its users. We express a preference to maintain consistency in development by using a single programming language, and for it to be affordable or an open source project. We describe whether and how these aspects are realized in each competing solution.

The reviewed simulations are divided into two main categories: (1) software platforms targeting simulation on CPUs, GPUs, or both; (2) hardware implementations of spiking neural networks, as described below.

\subsection{Software Simulation}
\label{ss:software_review}

Many spiking neural network frameworks exist, each with a unique set of use cases. Some focus on the biologically realistic simulation of neurons, while others on high-level spiking network functionality. To build a network to run even the simplest machine learning experiment, one will face multiple difficult design choices: Which biological properties should the neurons and the network have? E.g., how many GABAergic neurons or NMDA / AMPA receptors to use, or what form of synapse dynamics? Many such options exist, some of which may or may not have a significant impact on the performance of an ML system.

For example, \texttt{NEURON} \citep{NEURON}, \texttt{Genesis} \citep{GENESIS3.0}, \texttt{NEST} \citep{Gewaltig_NEST}, \texttt{BRIAN} \citep{Stimberg2014EquationorientedSO}, and \texttt{ANNarchy} \citep{Vitay2015ANNarchyAC} focus on accurate biological realistic simulation from sub-cellular components and biochemical reactions to complex models of single neurons up to the network level. These simulation platforms target the neuro-biophysics community and neuroscientists that wish to simulate multicompartment  neurons, where each sub-compartment is a different part of the neuron with different functionalities, morphological details, and shape. These packages are able to simulate large SNNs on various types of systems, from laptops all the way up to HPC systems. However, each simulated component must be \emph{homogeneous}, meaning that it must be built with a single type of neuron and a single type of synapse. If a researcher wants to simulate multiple types of neurons utilizing various synapse types, it may be difficult in these frameworks. For a more detailed comparison of development time, model performance, and varieties of models of neurons available in this biophysical software see \citep{ComparativeSoftwareforBrainNetworkSimulations}.

A major benefit of the \texttt{BRIAN}, \texttt{ANNarchy}, and \texttt{NEST} packages is that, besides the built-in modules for neuron and connection objects, the programmer is able to specify the dynamics of neurons and connections using differential equations. This eliminates the need to manually specify the dynamic properties of each new neuron or connection object in code. The equations are compiled into fast \texttt{C++} code in case of \texttt{ANNarchy}, vectorised and linear algebraic operations using NumPy and Basic Linear Algebra Subprograms (BLAS) in case of \texttt{BRIAN} and to a mix of Python and native “C-like” language (hoc) \citep{NEURONandPython} which are responsible for the SNN simulation. In addition, in the \texttt{NEST} package, the programmer can combine pre-configured objects (which accepts arguments) to create SNNs. In all of these libraries, significant changes to the operation of the network components requires modification of the underlying code, a difficult task which stands in the way of fast network prototyping and break the continuity of the programming. At this time \texttt{BindsNET} does not support the solving of arbitrary differential equations describing neural dynamics, but rather, for simplicity, several popular neuron types are supported.

Frameworks such as \texttt{NeuCube} \citep{KASABOV201462} and \texttt{Nengo} \citep{bekolay2014} focus on high-level behaviors of neural networks and may be used for machine learning experimentation. \texttt{Neucube} supports rate coding-based spiking networks, and \texttt{Nengo} supports simulation at the level of spikes, firing rates, or high-level, abstract neural behavior. \texttt{NeuCube} attempts to map spatiotemporal input data into three-dimensional SNN architectures; however, it is not an open source project, and therefore is somewhat restricted in scope and usability. \texttt{Nengo} is often used to simulate high-level functionality of brains or brain regions, as a cognitive modeling toolbox implementing the Neural Engineering Framework \citep{Stewart2012ATO} rather than a machine learning framework. \texttt{Nengo} is an open source project, written in Python, and supports a \texttt{Tensorflow} \citep{tensorflow2015-whitepaper} backend to improve simulation speed and exploit some limited ML functionality. It also has options for deploying neural models on dedicated hardware platforms; e.g., SpiNNaker \citep{Plana2011SpiNNakerDA}.

Other frameworks, e.g., \texttt{CARLsim} \citep{Beyeler2015CARLsim3A} and \texttt{NeMo} \citep{Fidjeland2009NeMoAP}, focus on the high-level aspects of spiking neural networks, and are more suited to machine learning applications. Both packages allow simulation of large spiking networks built from Izhikevich neurons \citep{Izhikevich2003SimpleMO} with realistic synaptic dynamics as their fundamental building block, and accelerate computation using GPU hardware. Like the frameworks before, low-level simulation code is written in \texttt{C++} for efficiency, and programmers may interface with them with the simulator-independent \texttt{PyNN} Python library \citep{Davison2008PyNNAC}, or in \texttt{MATLAB} or \texttt{Java}.

The above packages are typically written in more than one programming language: the core functionality is implemented in a lower-level language (e.g., \texttt{C++}) to achieve good performance, and the code exposed to the user of the package is written in a higher-level language (e.g., Python or MATLAB) to enable fast prototyping. If such frameworks are not tailored to the needs of a user, lead to steep learning curves, and aren't flexible enough to create a desired model, the user may have to program in both high- and low-level languages to make changes to the required internal components.

\subsection{Neuromorphic Hardware Simulation}
\label{ss:hardware_review}

Spiking neural network frameworks based on hardware have several advantages over software simulation in terms of performance and power consumption. Different hardware solutions have different use cases. For example, SpiNNaker \citep{Plana2011SpiNNakerDA} combines cheap, generic, yet dedicated CPU boards together to create a powerful SNN simulation framework in hardware. Other platforms (e.g., TrueNorth \citep{Akopyan2015TrueNorthDA}, HRL, and Braindrop) involve the design of a new chip, used to emulate spiking neurons by making use of the physical properties of certain materials. A novel development is Intel's Loihi platform for spike-based computation, outperforming all known conventional solutions \citep{davies2018loihi}. Other solutions are based on programmable hardware, like FPGAs which transform neural equations to configurations of electronic gates in order to speed up computation. More specialized hardware such as ASIC and DSP can be used to parallelize and therefore accelerate the calculations required for the operation and interaction of neurons in a spiking network.

In order to conduct experiments in the hardware domain, one must usually learn a specific programming language targeted to the hardware platform, or carefully adapt an existing experiment to the unique hardware environment under the constraints as enforced by chip designers; e.g., the connectivity limitations of the TrueNorth chip, or the limited configuration options of the SpiNNaker platform. In either case, this is not an ideal situation for researchers who want to rapidly prototype and test the capabilities of SNNs in simulation, machine learning, or other domains.

\section{Package structure}
\label{s:structure}

\begin{figure}[H]
    \centering
    \caption{Depiction of the \texttt{BindsNET} directory structure and description of software modules.}
    \includegraphics[width=1.0\textwidth]{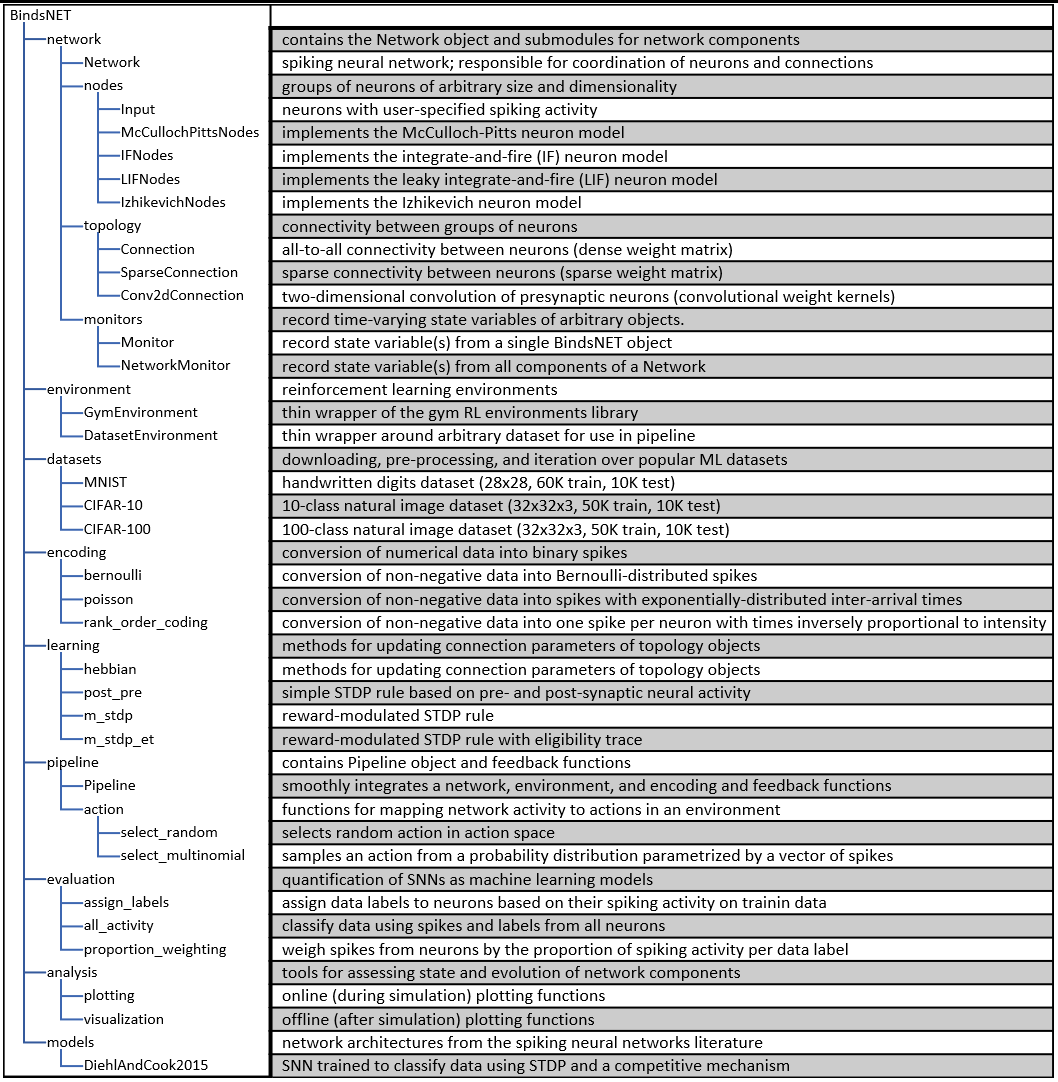}
    \label{fig:bindsnet_tree}
\end{figure}

A summary of all the software modules of the \texttt{BindsNET} package is included in Figure \ref{fig:bindsnet_tree}.

Many \texttt{BindsNET} objects use the \texttt{torch.Tensor} data structure for computation; e.g., all objects supporting the \texttt{Nodes} interface use \texttt{Tensor}s to store and update state variables such as spike occurrences or voltages. The \texttt{Tensor} object is a multi-dimensional matrix containing elements of a single data type; e.g., integers or floating points numbers with 8, 16, 32, or 64 bits of precision. They can be easily moved between devices with calls to \texttt{Tensor.cpu()} or \texttt{Tensor.gpu()}, and can target GPU devices by default with a call like \texttt{torch.set\_default\_tensor\_type('torch.cuda.FloatTensor')}.

\subsection{SNN simulation}
\label{ss:snn_simulation}

\texttt{BindsNET} provides a \texttt{Network} object (in the \texttt{network} module) which is responsible for the coordination of one or many \texttt{Nodes} and \texttt{Connections} objects, and supports the use of \texttt{Monitor}s for recording the state variables of these components. A time step parameter \texttt{dt} is the sole (optional) argument to the \texttt{Network} constructor, to which all other objects' time constants are scaled. The \texttt{run(inpts, time)} function implements synchronous updates (for a number of time steps $\frac{\texttt{time}}{\texttt{dt}}$) of all network components. This function calls \texttt{get\_inputs()} to calculate presynaptic inputs to all \texttt{Nodes} instances (alongside user-defined inputs in \texttt{inpts}) as a subroutine. A \texttt{\_reset()} method invokes resetting functionality of all network components, namely for resetting state variables back to default values. Saving and loading of networks to and from disk is implemented, permitting re-use of trained connection weights or other parameters.

The \texttt{Nodes} abstract base class in the \texttt{nodes} module specifies the abstract functions \texttt{step(inpts, dt)} and \texttt{\_reset()}. The first is called by the \texttt{run()} function of a \texttt{Network} instance to carry out a single time step's update, and the second resets spikes, voltages, and any other recorded state variables to default values. Implementations of the \texttt{Nodes} class include \texttt{Input} (neurons with user-specified or fixed spikes) \texttt{McCullochPittsNodes} (McCulloch-Pitts neurons), \texttt{IFNodes} (integrate-and-fire neurons), \texttt{LIFNodes} (leaky integrate-and-fire neurons), and \texttt{IzhikevichNodes} (Izhikevich neurons). Other neurons or neuron-like computing elements may be implemented by supporting the \texttt{Nodes} interface. Many \texttt{Nodes} object support optional arguments for customizing neural attributes such as threshold, reset, and resting potential, refractory period, membrane time constant, and more.

The \texttt{topology} module is used to specify interactions between \texttt{Nodes} instances, the most generic of which is implemented in the \texttt{Connection} object. The \texttt{Connection} is aware of \emph{source} (presynaptic) and \emph{target} (postsynaptic) \texttt{Nodes}, as well as a matrix of weights \texttt{w} of connections strengths. By default, connections do not implement any learning of connection weights, but do so through the inclusion of an \texttt{update\_rule} argument. Several canonical learning rules from the biological learning literature are implemented in the \texttt{learning} module, including Hebbian learning (\texttt{hebbian()}), a variant of spike-timing-dependent plasticity (STDP) (\texttt{post\_pre()}), and less well-known methods such as reward-modulated STDP (\texttt{m\_stdp()}). The optional argument \texttt{norm} to the \texttt{Connection} specified a desired sum of weights per target neuron, which is enforced by the parent \texttt{Network} during each call of \texttt{run()}. A \texttt{SparseConnection} object is available for specifying connections where certain weights are fixed to zero; however, this does not (yet) support learning functions. The \texttt{Conv2dConnection} object implements a two-dimensional convolution operation (using \texttt{PyTorch}'s \texttt{torch.nn.conv2d} function) and supports all update rules from the \texttt{learning} module.

\subsection{Machine and reinforcement learning}
\label{ss:ml_and_rl}

\texttt{BindsNET} is being developed with machine and reinforcement learning applications in mind. At the core of these efforts is the \texttt{learning} module, which contains functions which can be attached to \texttt{Connection} objects to modify them during SNN simulation. By default, connections are instantiated with no learning rule. The \texttt{hebbian} rule (``fire together, wire together'') symmetrically strengthens weights when pre- and postsynpatic spikes occur temporally close together, and the \texttt{post\_pre} rule implements a simple form of STDP in which weights are increased or decreased according to the relative timing of pre- and postsynaptic spikes, with user-specified (possibly asymmetric) learning rates. The reward-modulated STDP (\texttt{m\_stdp}) and reward-modulated STDP with eligibility trace (\texttt{m\_stdp\_et}) rules of \citet{Florian2007ReinforcementLT} are also implemented for use in basic reinforcement learning experiments.

The \texttt{datasets} module provides a means to download, pre-process, and iterate over machine learning datasets. For example, the \texttt{MNIST} object provides this functionality for the MNIST handwritten digits dataset. Several other datasets are supported besides, including CIFAR-10, CIFAR-100, and Spoken MNIST. The samples from a dataset can be encoded into spike trains using the \texttt{encoding} module, currently supporting several functions for creating spike trains from non-negative data based on different statistical distributions and biologically inspired transformations of stimuli. Spikes may be used as input to SNNs, or even to other ML systems. A submodule \texttt{preprocess} of \texttt{datasets} allows the user to apply various pre-processing techiques to raw data; e.g., cropping, subsampling, binarizing, and more.

The \texttt{environment} module provides an interface into which a SNN, considered as a reinforcement learning agent, can take input from and enact actions in an reinforcement learning environment. The \texttt{GymEnvironments} object comprises of a generic wrapper for \texttt{gym} \citep{BrockmanCPSSTZ16} RL environments and calls its \texttt{reset()}, \texttt{step(action)}, \texttt{close()}, and \texttt{render()} functionality, while providing a default pre-processing function \texttt{preprocess()} for observations from each environment. The \texttt{step(action)} function takes an \texttt{action} in the \texttt{gym} environment, which returns an observation, reward value, an indication of whether the episode has finished, and a dictionary of (name, value) pairs containing additional information. Another object, \texttt{DatasetEnvironment}, provides a generic wrapper around objects from the \texttt{datasets} module, allowing these to be used as a component in a \texttt{Pipeline} instance (see Section \ref{ss:pipeline}). The \texttt{environment.action} module provides methods for mapping one or more network layers' spikes to actions in the environment; e.g., \texttt{select\_multinomial()} treats a (normalized) vector of spikes as a probability distribution from which to sample an action for the environment's similarly-sized action space.

Simple methods for the evaluation of SNNs as machine learning models are implemented in the \texttt{evaluation} module. In the context of unsupervised learning, the \texttt{assign\_labels()} function assigns data labels to neurons corresponding to the class of data on which they spike most during network training \citep{p._u._diehl_unsupervised_2015}. These labels are to classify new data using methods like \texttt{all\_activity()} and \texttt{proportion\_weighting()} \citep{hazanetal2018}.

A collection of network architectures is defined in the \texttt{models} module. For example, the network structure of \citet{p._u._diehl_unsupervised_2015} is implemented by the \texttt{DiehlAndCook2015} object, which supports arguments such as \texttt{n\_neurons}, \texttt{excite}, \texttt{inhib}, etc. with reasonable default values.

\subsection{The Pipeline object}
\label{ss:pipeline}

As an additional effort to ease prototyping of machine learning systems comprising spiking neural networks, we have provided the \texttt{Pipeline} object to compose an environment, network, an encoding of environment observations, and a mapping from network activity to the environment's action space. The \texttt{Pipeline} also provides optional arguments for visualization of the environment and network state variables during network operation, skipping or recording observations on a regular basis, the length of the simulation per observation (defaults to 1 time step), and more. The main action of the pipeline can be explained as a four-step, recurring process, implemented in the pipeline \texttt{step()} function:

\begin{enumerate}
    \item An action is selected based on the activity of one or more of the network's layers during the last one or more time steps
    \item This action is used as input to the environment's \texttt{step()} function, which returns a new observation, a scalar reward, whether the simulation has finished, and any additional information particular to the environment
    \item The observation returned from the environment is converted into spike trains according to the user-specified encoding function (either custom or from the \texttt{encoding} module) and request simulation time
    \item The spike train-encoded observation is used as input to the network
\end{enumerate}

Alongside the required arguments for the \texttt{Pipeline} object (\texttt{network}, \texttt{environment}, \texttt{encoding}, and \texttt{action}), there are a few keyword arguments that are supported, such as \texttt{history} and \texttt{delta}. The \texttt{history\_length} argument indicates that a number of sequential observations are to maintained in order to calculate differences between current observations and those stored in the \texttt{history} data structure. This implies that only new information in the environment's observation space is delivered as input to the network on each time step. The \texttt{delta} argument (default 1) specifies an interval at which observations are stored in \texttt{history}. This may be useful if observations don't change much between consecutive steps; then, we should wait some \texttt{delta} time steps between taking observations to expect significant differences. As an example, combining \texttt{history\_length = 4} and \texttt{delta = 3} will store observations \{0, 3, 6, 9\}, \{3, 6, 9, 12\}, \{6, 9, 12, 15\}, etc. A few other keyword arguments for handling console output, plotting, and more exist and are detailed in the \texttt{Pipeline} object documentation.

\begin{figure}[H]
    \centering
    \includegraphics[width=0.6\textwidth]{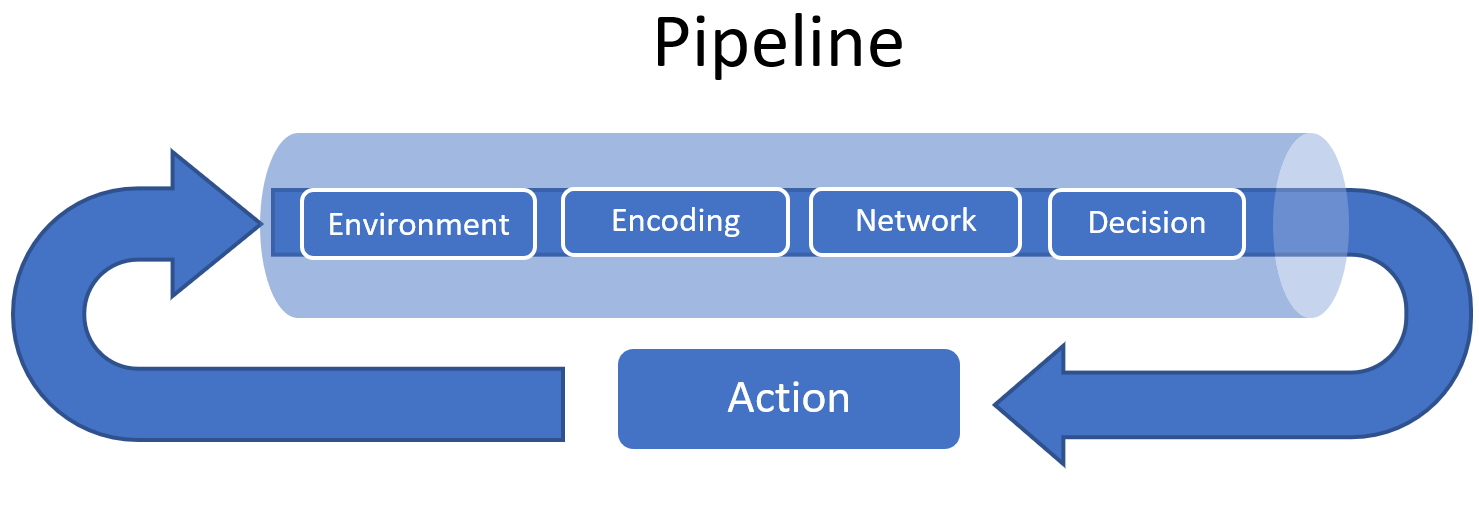}
    \caption{A functional diagram of the \texttt{Pipeline} object. The four-step process involves an encoding function, network computation, converting network outputs into actions in an environment's action space, and a simulation step of the environment. An encoding function converts non-spiking observations from the environment into spike inputs to the network, and a \texttt{action} function maps network spiking activity into a non-spiking quantity: an action, fed back into the environment, where the procedure begins anew. Other modules come into play in various supporting roles: the network may use a \texttt{learning} method to update connection weights, or the environment may simply be a thin wrapper around a dataset (in which case there is no feedback), and it may be desirable to plot network state variables during the reinforcement learning loop.}
    \label{fig:pipeline}
\end{figure}

A functional diagram of the \texttt{Pipeline} object is depicted in Figure \ref{fig:pipeline}.

\subsection{Visualization}
\label{ss:visualization}

\texttt{BindsNET} also contains useful visualization tools that provide information during network or environment simulation. Several generic plotting functions are implemented in the \texttt{analysis.plotting} module; e.g., \texttt{plot\_spikes()} and \texttt{plot\_voltages()} create and update plots dynamically instead of recreating figures at every time step. These functions are able to display spikes and voltages with a single call. Other functions include \texttt{plot\_weights()} (displays connection weights), \texttt{plot\_input()} (displays raw input data), and \texttt{plot\_performance()} (displays time series of performance metric).

The \texttt{analysis.visualization} module contains additional plotting functionality for network state variables after simulation has finished. These tools should allow experimenters to analyzed learned weights or spike outputs, or to summarize long-term behaviors of their SNN models. For example, the \texttt{weights\_movie()} function creates an animation of a \texttt{Connection}'s weight matrix from a sequence of its values, enabling the visualization of the trajectory of connection weight updates.

\section{Examples}
\label{s:examples}

We present some simple example scripts to give an impression of how \texttt{BindsNET} can be used to build spiking neural networks implementing machine learning functionality. \texttt{BindsNET} is built with the concept of encapsulation of functionality to make it faster and easier for generalization and prototyping. Note in the examples below the compactness of the scripts: fewer lines of code are needed to create a model, load a dataset, specify their interaction in terms of a pipeline, and run a training loop. Of course, these commands rely on many lines of underlying code, but the user no longer has to implement them for each experimental script. If changes in the available parameters are not enough, the experimenter can intervene by making changes in the underlying code in the model without changing language or environment, thus preserving the continuity of the coding environment.

\subsection{Unsupervised learning}
\label{ss:unsupervised}

\begin{figure}[H]
    \includegraphics[width=\textwidth]{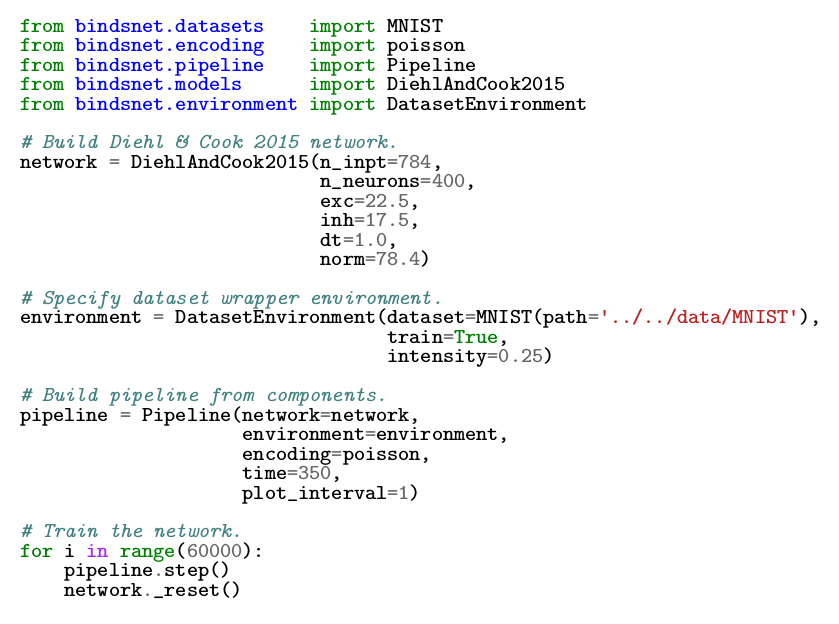}
    \caption{Unsupervised learning of the MNIST handwritten digits in \texttt{BindsNET}. The \texttt{DiehlAndCook2015} model implements a simple spike timing-dependent plasticity rule between input and excitatory neuron populations as well as a competitive inhibition mechanism to learn prototypical digit filters from raw data. The \texttt{DatasetEnvironment} wraps the \texttt{MNIST} dataset object so it may be used as a component in the \texttt{Pipeline}. The network is trained on one pass through the 60K-example training data for 350ms each, with state variables (voltages and spikes) reset after each example.}
    \label{code:unsupervised_mnist}
\end{figure}

\begin{figure}[H]
  \centering
  \begin{subfigure}[t]{5cm}
    \includegraphics[height=3.5cm, width=5cm]{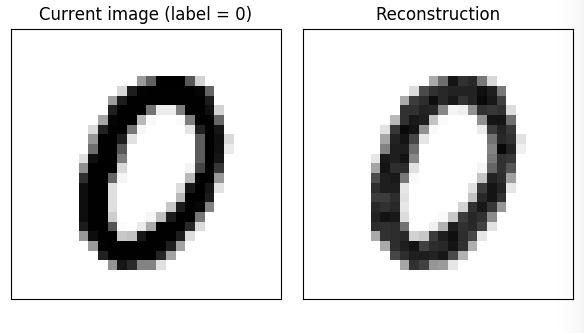}
    \caption{Raw input and ``reconstructed'' input, computed by summing Poisson-distributed spike trains over the time dimension.}
  \end{subfigure}
  \quad
  \begin{subfigure}[t]{5cm}
    \includegraphics[height=3.5cm, width=5cm]{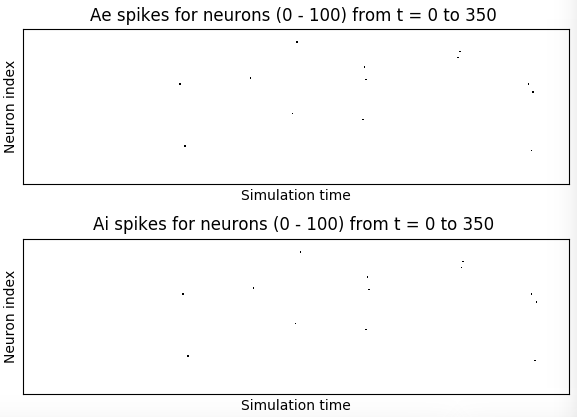}
    \caption{Spikes from the excitatory and inhibitory layers of the \texttt{DiehlAndCook2015} model.}
  \end{subfigure}
  \quad
  \begin{subfigure}[t]{5cm}
    \includegraphics[height=3.5cm, width=5cm]{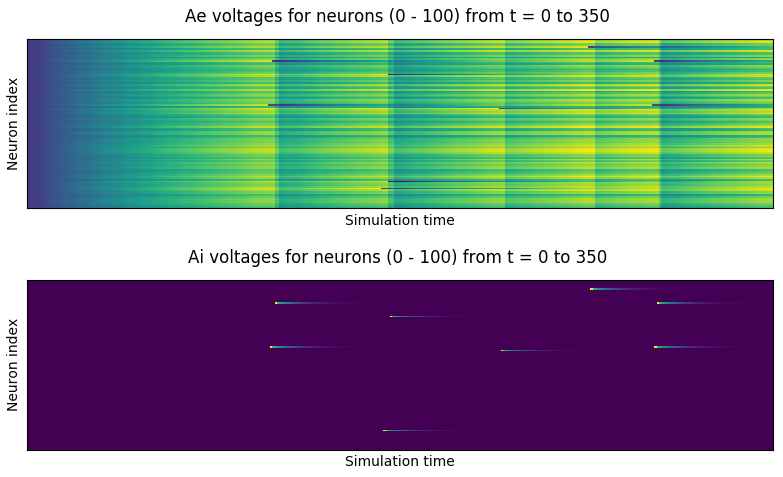}
    \caption{Voltages from the excitatory and inhibitory layers of the \texttt{DiehlAndCook2015} model.}
  \end{subfigure}
  
  \medskip
  
  \begin{subfigure}[t]{5cm}
    \includegraphics[height=5cm, width=5cm]{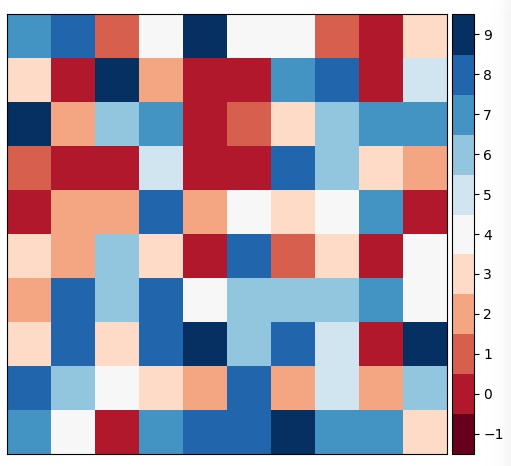}
    \caption{Reshaped 2D label assignments of excitatory neurons, assigned based on activity on examples from the training data.}
  \end{subfigure}
  \quad
  \begin{subfigure}[t]{5cm}
    \includegraphics[height=5cm, width=5cm]{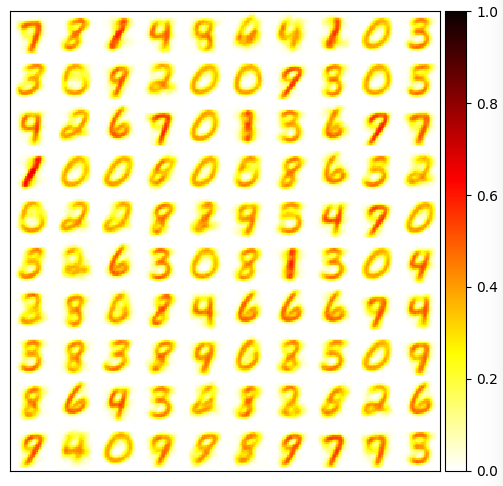}
    \caption{Reshaped 2D connection weights from input to excitatory layers. The network is able to learn distinct prototypical examples from the dataset, corresponding to the categories in the data.}
  \end{subfigure}
  \caption{Accompanying plots to the unsupervised training of the \texttt{DiehlAndCook2015} spiking neural network architecture. The network is able to learn prototypical examples of images from the training set, and on a test images, the excitatory neuron with the most similar filter should fire first.}
  \label{fig:unsupervised_mnist}
\end{figure}

The \texttt{DiehlAndCook2015} object in the \texttt{models} module implements a slightly simplified version of the network architecture discussed in \citet{p._u._diehl_unsupervised_2015}. A minimal working example of training a spiking neural network to learn, without labels, a representation of the MNIST digit dataset is given in Figure \ref{fig:unsupervised_mnist}, and state variable-monitoring plots are depicted in Figure \ref{code:unsupervised_mnist}. The \texttt{Pipeline} object is used to hide the messy details of the coordination between the dataset, encoding function, and network instance.  Code for additional plots or console output may be added to the training loop for monitoring purposes as needed.

\subsection{Supervised learning}
\label{ss:supervised}

\begin{figure}[H]
    \centering
    \includegraphics[width=\textwidth]{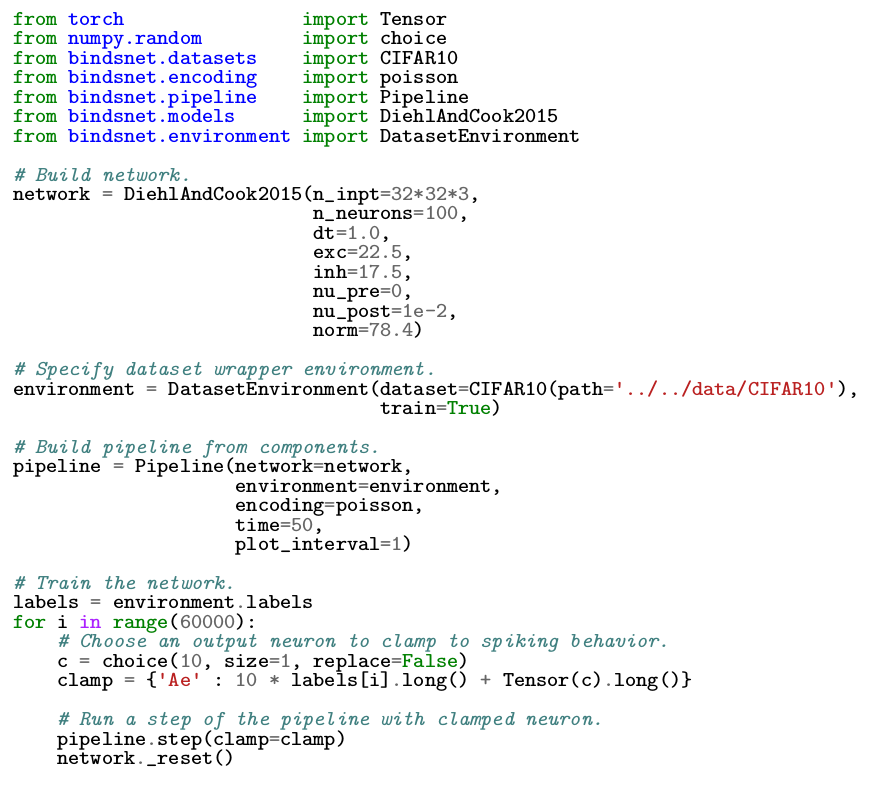}
    \caption{The \texttt{DiehlAndCook2015} model is instantiated with 100 neurons, and is otherwise identical to the model in the unsupervised learning example. The group of neurons named ``Ae'' refers to the excitatory layer of neurons in the \citet{p._u._diehl_unsupervised_2015} model. A random neuron from the subgroup corresponding to the data label is selected, and clamped to spike on each time step (using \texttt{pipeline.step(clamp=clamp)}).}
    \label{code:supervised_cifar10}
\end{figure}

\begin{figure}[H]
  \centering
  \begin{subfigure}[t]{5cm}
    \includegraphics[height=3.5cm, width=5cm]{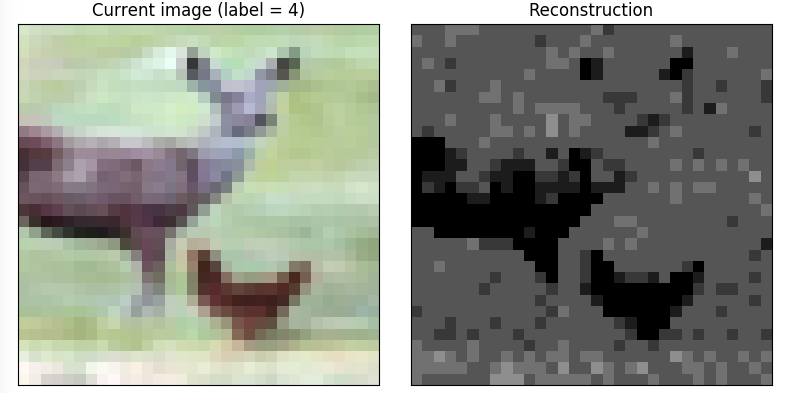}
    \caption{Raw input and ``reconstructed'' input, computed by summing Poisson-distributed spike trains over the time dimension.}
  \end{subfigure}
  \quad
  \begin{subfigure}[t]{5cm}
    \includegraphics[height=3.5cm, width=5cm]{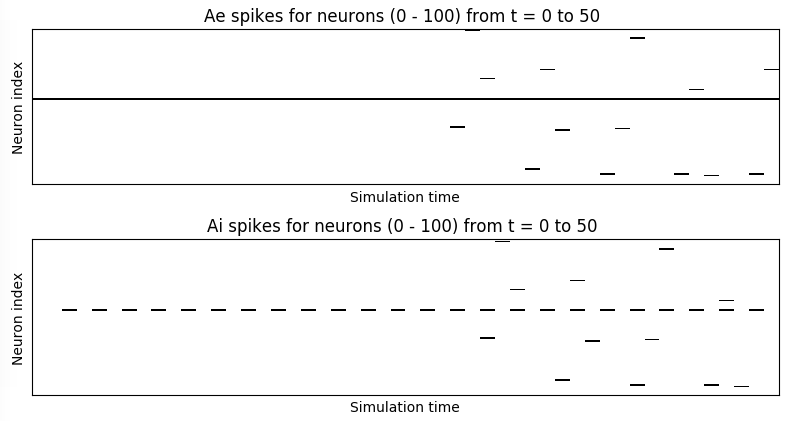}
    \caption{Spikes from the excitatory and inhibitory layers of the \texttt{DiehlAndCook2015} model.}
  \end{subfigure}
  \quad
  \begin{subfigure}[t]{5cm}
    \includegraphics[height=3.5cm, width=5cm]{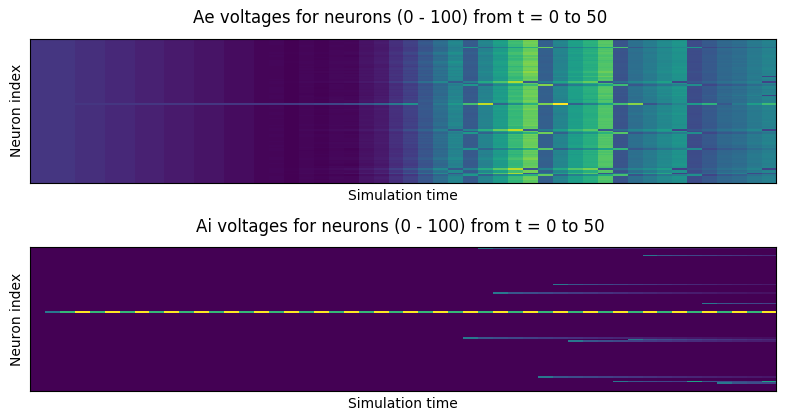}
    \caption{Voltages from the excitatory and inhibitory layers of the \texttt{DiehlAndCook2015} model.}
  \end{subfigure}
  
  \medskip
  
  \begin{subfigure}[t]{5cm}
    \includegraphics[height=5cm, width=5cm]{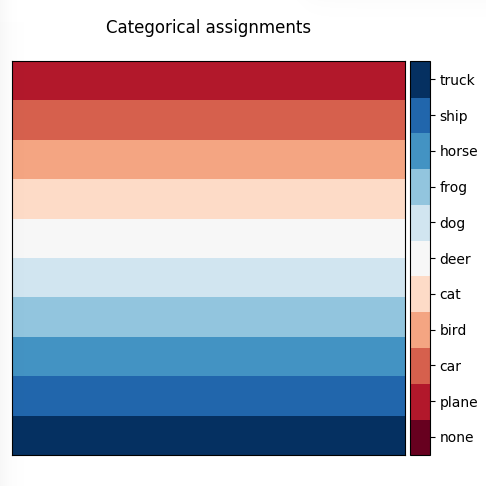}
    \caption{Reshaped two-dimensional label assignments of excitatory neurons, assigned based on activity on examples from the training data.}
  \end{subfigure}
  \quad
  \begin{subfigure}[t]{5cm}
    \includegraphics[height=5cm, width=5cm]{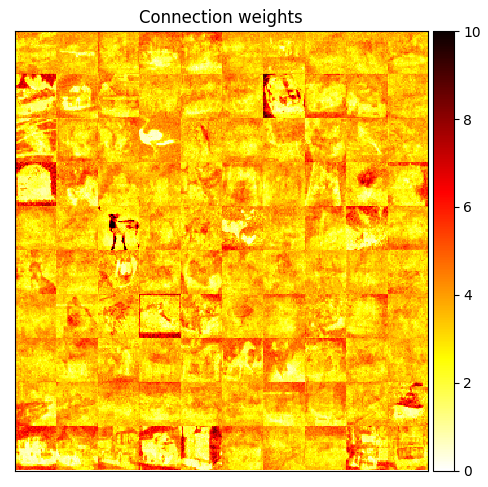}
    \caption{Reshaped two-dimensional weights connected input to excitatory layers. Certain groups of neurons are forced to represent a certain digit class.}
  \end{subfigure}
  \caption{Accompanying plots for supervised training of a \texttt{DiehlAndCook2015} network with 400 excitatory and inhibitory neurons, trained by clamping random neurons from a particular subpopulation to spike when a particular input category is presented. The first twenty neurons represent the image category ``plane'', the second twenty neurons represent the image category ``car'', and so on. STDP is used to update the synapse weights.}
  \label{fig:supervised_cifar10}
\end{figure}

We can use a modified version of the \texttt{DiehlAndCook2015} network to implement supervised learning of the CIFAR-10 natural images dataset. An minimal example of training a spiking network to classify the CIFAR-10 dataset is given in Figure \ref{code:supervised_cifar10}, with plotting outputs depicted in Figure \ref{fig:supervised_cifar10}. A layer of 100 excitatory neurons is split into 10 groups of size 10, one for each digit category. On each input example, we observe the label of the data and clamp a randomly selected excitatory neuron from its group to spike on every time step. This forces the neuron to adjust its filter weights towards the shape of current input example.

\subsection{Reinforcement learning}
\label{ss:reinforcement}

\begin{figure}[H]
    \centering
    \includegraphics[width=\textwidth]{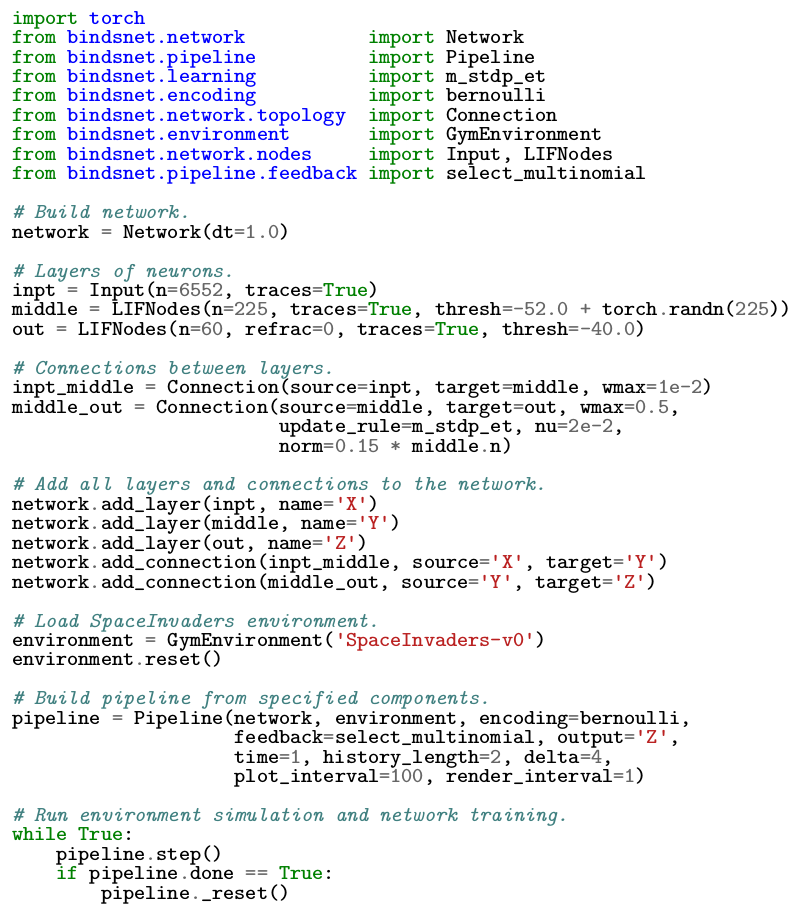}
    \caption{A spiking neural network that accept input from the \texttt{SpaceInvaders-v0} \texttt{gym} Atari environment. The observations from the environment are downsampled and binarized before they are converted into Bernoulli-distributed vectors of spikes, one per time step. The \texttt{history} and \texttt{delta} keyword arguments are used to create a reasonable level of input spiking activity. The output layer of the network has 60 neurons in it, each subsequent 10 representing a different action in the Space Invaders game as a population code. An action is selected at each time step using the \texttt{select\_multinomial} feedback function, which treats the averaged (normalized) activity over each output layer subpopulation as a probability distribution over actions.}
    \label{code:space_invaders}
\end{figure}

\begin{figure}[H]
  \centering
  \begin{subfigure}[t]{5cm}
    \includegraphics[height=3.5cm, width=5cm]{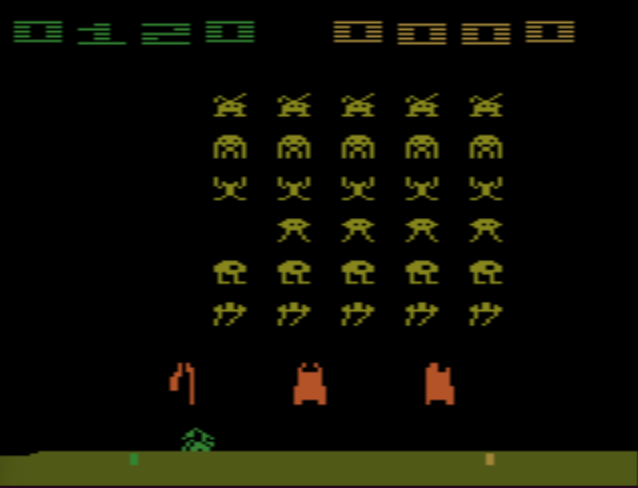}
    \caption{Raw output from the Space Invaders game, provided by the OpenAI \texttt{gym} \texttt{render()} method.}
  \end{subfigure}
  \qquad
  \begin{subfigure}[t]{5cm}
    \includegraphics[height=3.5cm, width=5cm]{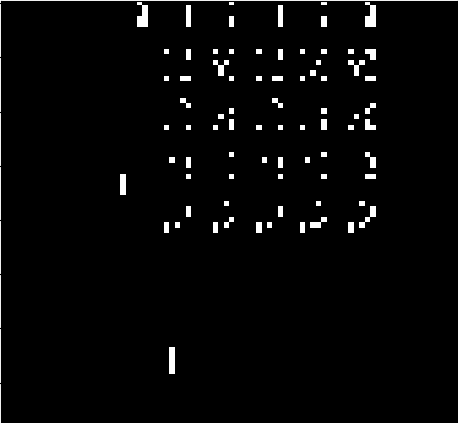}
    \caption{Pre-processed output from Space Invaders game environment used as input to the SNN.}
  \end{subfigure}
  
  \medskip
  
  \begin{subfigure}[t]{5cm}
    \includegraphics[height=3.5cm, width=5cm]{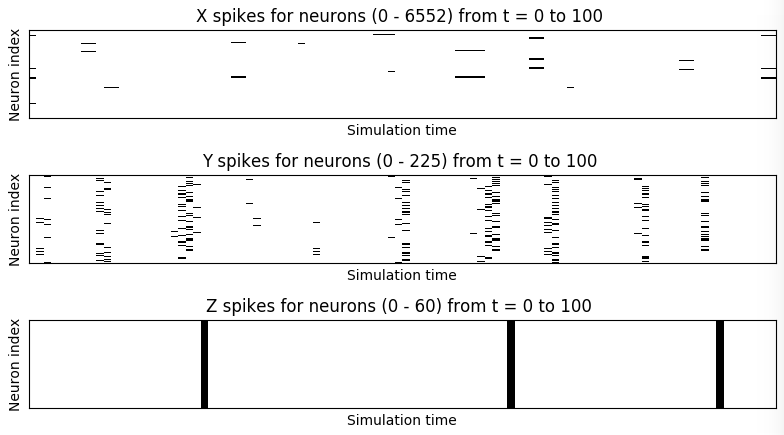}
    \caption{Spikes from the input, intermediate, and output layers of the spiking neural network.}
  \end{subfigure}
  \qquad
  \begin{subfigure}[t]{5cm}
    \includegraphics[height=3.5cm, width=5cm]{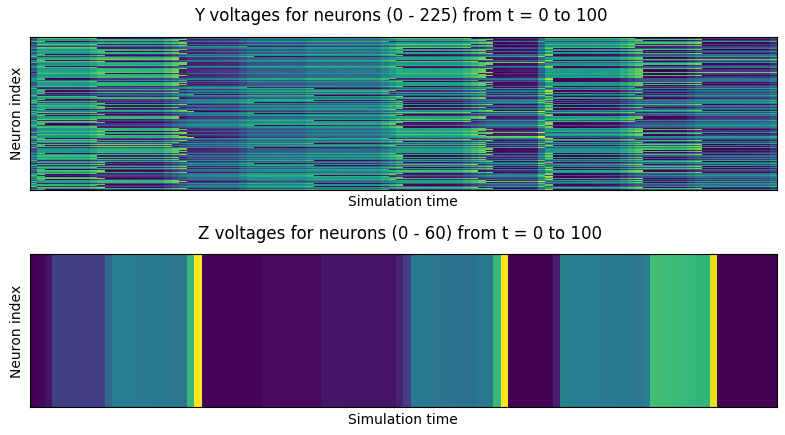}
    \caption{Voltages from the intermediate and output populations of the spiking neural network.}
  \end{subfigure}
  \caption{Accompanying plots for a custom spiking neural network's which interacts with the \texttt{SpaceInvaders-v0} reinforcement learning environment. Spikes and voltages of all neuron populations are plotted, and the Space Invaders game is rendered, as well as the downsampled, \texttt{history}- and \texttt{delta}-altered observation, which is presented to the network.}
  \label{fig:space_invaders}
\end{figure}

For a more complete view of the details involved in constructing an SNN and deploying a \texttt{GymEnvironment} instance, see the script depicted in Figure \ref{code:space_invaders} and accompanying displays in Figure \ref{fig:space_invaders}. A three layer SNN is built to compute on spikes encoded from Space Invaders observations. The result of this computation is some spiking activity in the \texttt{out} layer, which are converted into actions in the game's action space. The simulation of both the network and the environment are interleaved and appear to operate in parallel, and the network is able to adjust its parameters \emph{online} using the reward-modulated STDP update rule \texttt{m\_stdp\_et} taken from \citet{Florian2007ReinforcementLT}.

\subsection{Reservoir Computing}
\label{ss:reservoir}

\begin{figure}[H]
    \centering
    \includegraphics[width=\textwidth]{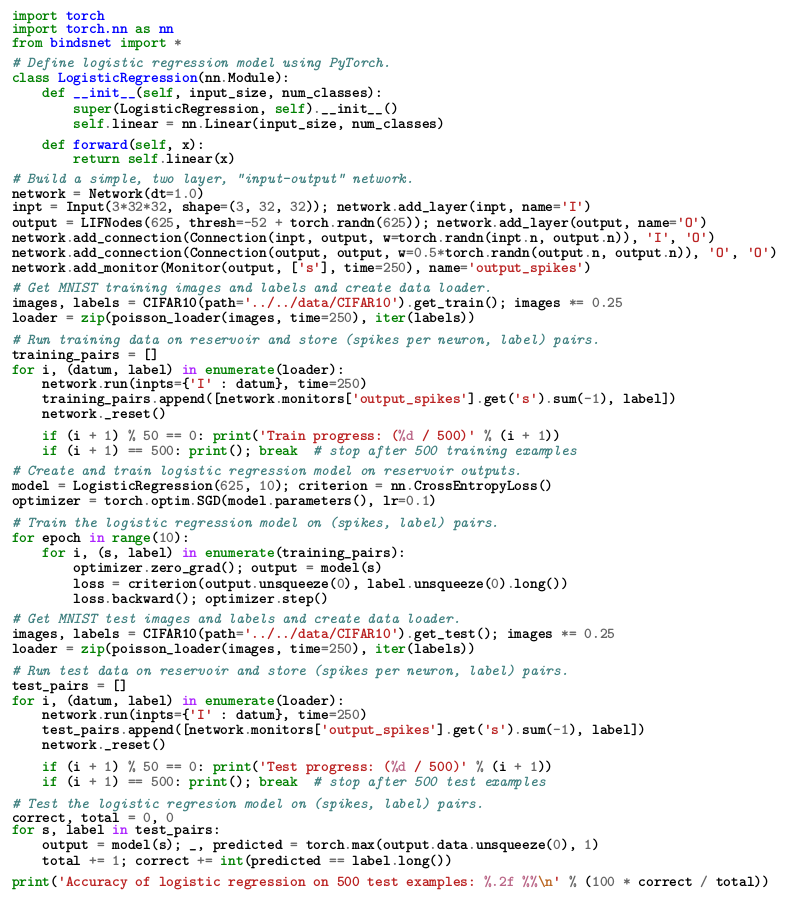}
    \caption{A recurrent neural network built from 625 spiking neurons accepts inputs from the CIFAR-10 natural images dataset. An \emph{input} population is connected all-to-all to an \emph{output} population of LIF neurons with weights draw from the standard normal distribution, which has voltage thresholds drawn from $\mathcal{N} (-52, 1)$ and is recurrently connected to itself with weights drawn from $\mathcal{N} (0, \frac{1}{2})$. The reservoir is used to create a high-dimensional, temporal representation of the image data, which is used to train and test a logistic regression model created with \texttt{PyTorch}.}
    \label{code:reservoir}
\end{figure}

\begin{figure}[H]
  \centering
  \begin{subfigure}[t]{6cm}
    \includegraphics[height=3.5cm, width=6cm]{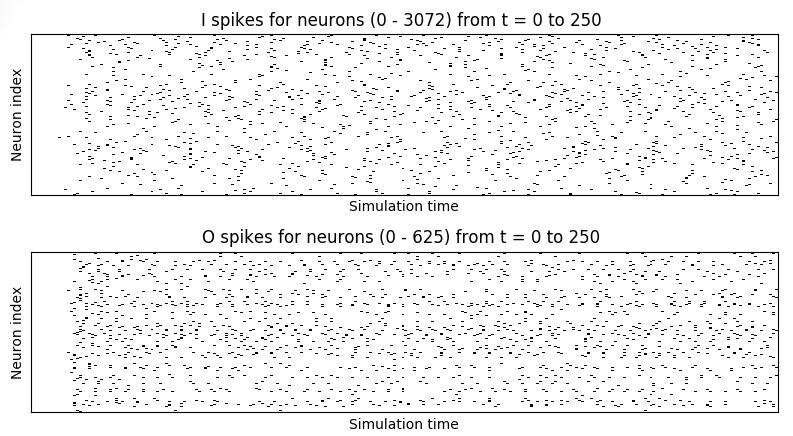}
    \caption{Spikes recorded from the input and output layers of the two layer reservoir network.}
  \end{subfigure}
  \quad\qquad
  \begin{subfigure}[t]{6cm}
    \includegraphics[height=3.5cm, width=6cm]{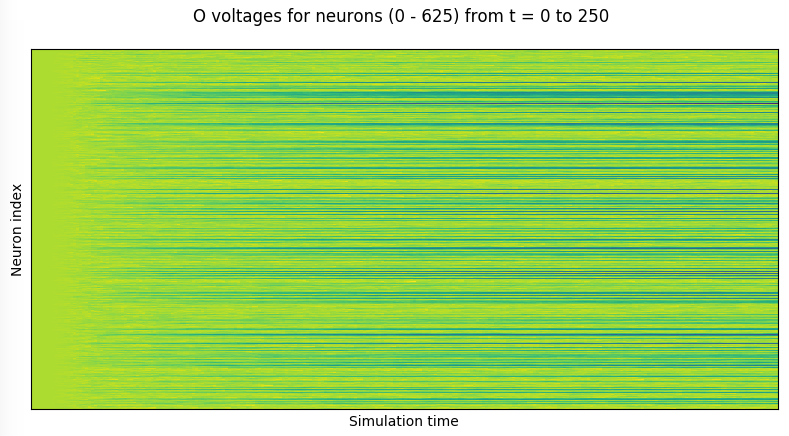}
    \caption{Voltages recorded from the output of the two layer reservoir network.}
  \end{subfigure}
  
  \medskip
  
  \begin{subfigure}[t]{6cm}
    \includegraphics[height=3cm, width=6cm]{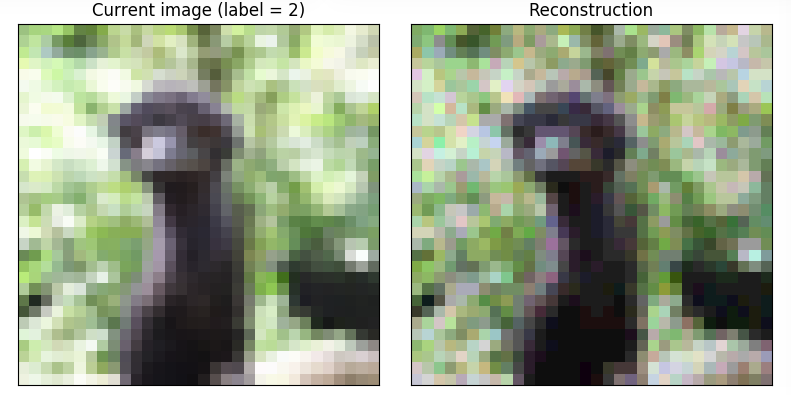}
    \caption{Raw input and its reconstruction, computed by summing Poisson-distributed spike trains over the time dimension.}
  \end{subfigure}
  \qquad
  \begin{subfigure}[t]{3cm}
    \includegraphics[height=3cm, width=3cm]{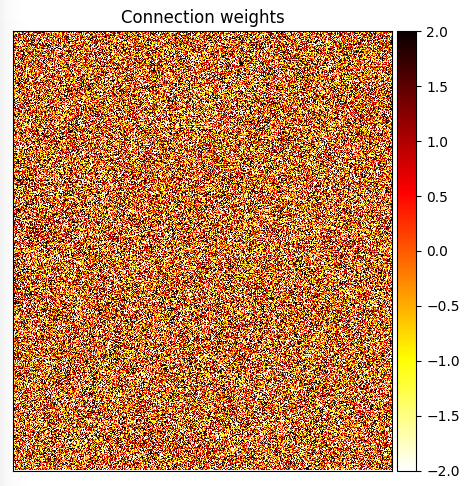}
    \caption{Weights from input to output neuron populations, initialized initialized from the distribution $\mathcal{N} (0, 1)$.}
  \end{subfigure}
  \quad
  \begin{subfigure}[t]{3cm}
    \includegraphics[height=3cm, width=3cm]{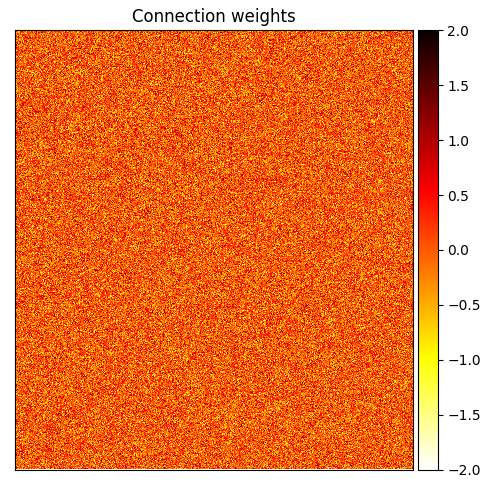}
    \caption{Recurrent weights of the output population, initialized from the distribution $\mathcal{N} (0, \frac{1}{2})$.}
  \end{subfigure}
  \caption{Plots accompanying another reservoir computing example, in which an input population of size equal to the CIFAR-10 data dimensionality is connected to a population of 625 LIF neurons, which is recurrently connected to itself.}
  \label{fig:cifar10_reservoir}
\end{figure}

Reservoir computers are typically built from three parts: (1) an encoder that translates input from the environment that is fed to it, (2) a dynamical system based on randomly connected neurons (the \textit{reservoir}), and (3) a readout mechanism. The readout is often trained via gradient descent to perform classification or regression on some target function. \texttt{BindsNET} can be used to build reservoir computers using spiking neurons with little effort, and machine learning functionality from \texttt{PyTorch} can be co-opted to learn a function from states of the high-dimensional reservoir to desired outputs. Code in for defining and simulating a simple reservoir computer is given in Figure \ref{code:reservoir}, and plots to monitor simulation progress are shown in Figure \ref{fig:cifar10_reservoir}. The outputs of the reservoir computer on the CIFAR-10 natural image dataset are used as transformed inputs to a logistic regression model. The logistic regression model is then trained to recognize the digits based on the features produced by the reservoir.

\section{Ongoing development}
\label{s:future}
\texttt{BindsNET} is still at an early stage of development, and there is much room for future work and improvement. Since it is an open source project and because there is considerable interest in the research community in using SNNs for machine learning purposes, we are optimistic that there will be numerous community contributions to the library. Indeed, we believe that public interest in the project, along with the strong support of the libraries on which it depends, will be an important driving factor in its maturation and proliferation of features. We mention some specific implementation goals:

\begin{itemize}
    \item Additional neuron types, learning rules, datasets, encoding functions, etc. Added features should take priority based on the needs of the users of the library.
    \item Specialization of machine learning and reinforcement learning algorithms for spiking neural networks. These may take the form of additional learning rules, or some more complicated training methods that operate at the network level rather than on individual synapses.
    \item Tighter integration with \texttt{PyTorch}. Much of \texttt{PyTorch}'s neural network functionality may have interpretations in the spiking neural network context (e.g., as with \texttt{Conv2dConnection}).
    \item Conversion from deep neural network models implemented in \texttt{PyTorch} or specified in the \texttt{ONNX} format to near-equivalent spiking neural networks (as in \citet{diehl_fast-classifying_2015}).
    \item Performance optimization: improving the performance of library primitives will save time on all experiments with spiking neural networks.
    \item Automatic smoothing of SNNs: approximating spiking neurons as differentiable operations will permit the application of backpropagation to \texttt{BindsNET} SNNs. The \texttt{torch.autograd} automatic differentiation library \citep{paszke2017automatic} can then easily be applied to optimize the parameters spiking networks for ML problems.
\end{itemize}

\section{Discussion}
\label{s:discussion}

We have presented the \texttt{BindsNET} open source package for rapid biologically inspired prototyping of spiking neural networks with a machine learning-oriented approach. \texttt{BindsNET} is developed entirely in Python and is built on top of other mature Python libraries that lend their power to utilize multi-CPU or multi-GPU hardware configurations. Specifically, the ML tools and powerful data structures of \texttt{PyTorch} are a central part of \texttt{BindsNET}'s operation. \texttt{BindsNET} may also interface with the \texttt{gym} library to connect spiking neural networks to reinforcement learning environments. In sum, \texttt{BindsNET} represents an additional and attractive alternative for the research community for the purpose of developing faster and more flexible tools for SNN experimentation.

\texttt{BindsNET} comprises a spiking neural network simulation framework that is easy to use, flexible, and efficient. Our library is set apart from other solutions by its ML and RL focus; complex details of the biological neuron are eschewed in favor of high-level functionality. Computationally inclined researchers may be familiar with the underlying \texttt{PyTorch} functions and syntax, and excited by the potential of the third generation of neural networks for ML problems, driving adoption in both ML and computational neuroscience communities. This combination of ML programming tools and neuroscientific ideas may facilitate the further integration of biological neural networks and machine learning. To date, spiking neural networks have not been widely applied in ML and RL problems; having a library aimed at such is a promising step towards exciting new lines of research.

Researchers interested in developing spiking neural networks for use in ML or RL applications will find that \texttt{BindsNET} is a powerful and easy tool to develop their ideas. To that end, the biological complexity of neural components has been kept to a minimum, and high-level, qualitative functionality has been emphasized. However, the experimenter still has access to and control over groups of neurons at the level of membrane potentials and spikes, and connections at the level of synapse strengths, constituting a relatively low level of abstraction. Even with such details included, it is straightforward to build large and flexible network structures and apply them to real data. We believe that the ease with which our framework allows researchers to reason about spiking neural networks as ML models, or as RL agents, will enable advancements in biologically plausible machine learning, or further fusion of ML with neuroscientific concepts.

Although \texttt{BindsNET} is similar in spirit to the \texttt{Nengo} \citep{bekolay2014} neural and brain modeling software in that both packages can utilize a deep learning library as a ``backend'' for computation, \texttt{Nengo} optionally uses \texttt{Tensorflow} in a limited fashion while \texttt{BindsNET} uses \texttt{PyTorch} by default, for all network simulation functionality (with the \texttt{torch.Tensor} object). Additionally, for users that prefer the flexibility and the imperative execution of \texttt{PyTorch}, \texttt{BindsNET} inherits these features and is developed with many of the same design principles in mind.

We see \texttt{BindsNET} as a simple yet attractive option for those looking to quickly build flexible SNN prototypes backed by an easy-to-use yet powerful deep learning library. It encourages the conception of spiking networks as machine learning models or reinforcement learning agents, and is perhaps the first of its kind to provide a seamless interface to RL environments. The library is supported by several mature and feature-full open source software projects, and experiences immediate benefits from their growth and efficiency improvements. Considered as an extension of the \texttt{PyTorch} library, \texttt{BindsNET} represents a natural progression from the second to third generation of neural network research and development.

\section*{Acknowledgements}

This work has been supported in part by Defense Advanced Research Project Agency Grant, DARPA/MTO HR0011-16-l-0006 and by National Science Foundation Grant NSF-CRCNS-DMS-13-11165.

\bibliographystyle{abbrvnat}
\bibliography{main.bib}

\end{document}